\documentclass[a4paper,conference]{IEEEtran}
\usepackage{times}
\usepackage{epsfig}
\usepackage{graphicx}
\usepackage{amsmath}
\usepackage{amssymb}
\usepackage{pdfpages}
\usepackage{enumitem}
\usepackage{listings}

\usepackage{multirow}
\usepackage{adjustbox}

\usepackage{subfig}
\usepackage{gensymb}

\usepackage{fancyhdr}

\usepackage[hidelinks]{hyperref}
\hypersetup{
  colorlinks   = true, 
  urlcolor     = blue, 
  citecolor   = green 
}

\ifCLASSINFOpdf
\else
\fi
\begin{document}
%
\title{Simple Multi-Resolution Representation Learning for Human Pose Estimation}

\author{\IEEEauthorblockN{Trung Q. Tran, Giang V. Nguyen, Daeyoung Kim}
\IEEEauthorblockA{School of Computing\\
KAIST\\
Daejeon, South Korea\\
Email: \{trungtq2019, dexter.nguyen7, kimd\}@kaist.ac.kr}
}


%


\maketitle

\thispagestyle{fancy}
\pagestyle{fancy}
\fancyhf{}
\lhead{2020 25th International Conference on Pattern Recognition (ICPR)}
\rhead{Milan, Italy, Jan 10-15, 2021}
\cfoot{\thepage}

\begin{abstract}
Human pose estimation - the process of recognizing human keypoints in a given image - is one of the most important tasks in computer vision and has a wide range of applications including movement diagnostics, surveillance, or self-driving vehicle. The accuracy of human keypoint prediction is increasingly improved thanks to the burgeoning development of deep learning. Most existing methods solved human pose estimation by generating heatmaps in which the $i$th heatmap indicates the location confidence of the $i$th keypoint. In this paper, we introduce novel network structures referred to as multi-resolution representation learning for human keypoint prediction. At different resolutions in the learning process, our networks branch off and use extra layers to learn heatmap generation. We firstly consider the architectures for generating the multi-resolution heatmaps after obtaining the lowest-resolution feature maps. Our second approach allows learning during the process of feature extraction in which the heatmaps are generated at each resolution of the feature extractor. The first and second approaches are referred to as multi-resolution heatmap learning and multi-resolution feature map learning respectively. Our architectures are simple yet effective, achieving good performance. We conducted experiments on two common benchmarks for human pose estimation: MS-COCO and MPII dataset. The code is made publicly available at https://github.com/tqtrunghnvn/SimMRPose.
\end{abstract}


%
\IEEEpeerreviewmaketitle

\section{Introduction}

Human pose estimation is one of the vital tasks in computer vision and has received a great deal of attention from researchers for the past few decades. From the spatial aspect, this problem is divided into 2D and 3D human pose estimation. Geometrically, the 3D human pose might be predicted through the respective 2D human pose combining with a 3D exemplar matching. 
This paper focuses on the deep learning approach for 2D human pose estimation which aims to localize human anatomical keypoints on the torso, face, arms, and legs. 

\begin{figure}[h!]
\centering
\includegraphics[width=3.1in]{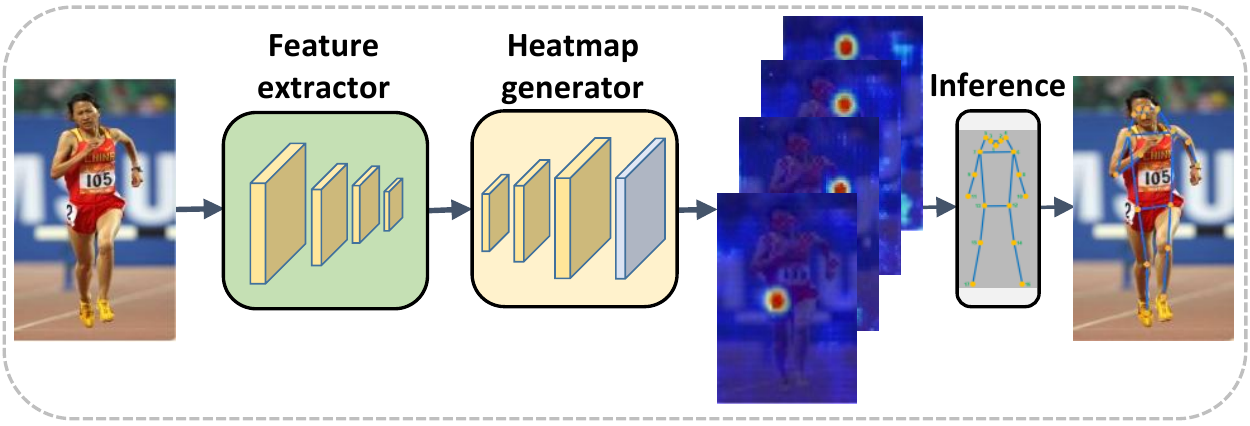}
\caption{Simple pipeline for human pose estimation using heatmaps.}
\label{fig_pipeline}
\end{figure}

The pioneer of deep learning methods formulated human pose estimation as a CNN-based regression towards body joints \cite{toshev2014deeppose}. The model uses an AlexNet 
backend (consisting of 7 layers) and an extra final layer that directly outputs joint coordinates. The later state-of-the-art methods reshaped this problem by estimating $k$ heatmaps for all $k$ human keypoints, where the $i$th heatmap represents the location confidence of the $i$th keypoint \cite{tompson2015efficient, wei2016convolutional, newell2016stacked, chen2018cascaded, xiao2018simple}. Heatmap-based approaches consist of two major parts as shown in Fig. \ref{fig_pipeline}: the first part (encoder) works as a feature extractor which is responsible for understanding the image while the second one (decoder) is to generate the heatmaps corresponding to the human keypoints. Convolutional pose machines (CPM) \cite{wei2016convolutional} used a multi-stage training scheme where the image features and the heatmaps produced by the previous stage are fed as the input; thus, the prediction is refined throughout stages. Commonly, the output of the feature extractor is the low-resolution feature maps. Stacked Hourglass \cite{newell2016stacked} and Cascaded pyramid network (CPN) \cite{chen2018cascaded} adopted a multi-resolution learning strategy to generate the heatmaps from the feature maps at a variety of resolutions. Instead of independently processing at multiple resolutions as CPN, Hourglass uses skip layers to preserve spatial information at each resolution. However, these two methods were defeated when Xiao et al. \cite{xiao2018simple} proposed a simple yet effective baseline which utilizes ResNet 
as its backbone for feature extractor followed by a few deconvolutional layers for heatmap generator (Fig. \ref{fig-pose-resnet}). SimpleBaseline \cite{xiao2018simple} for human pose estimation is the most effortless way to generate the heatmaps from the low-resolution feature maps, obtaining good performance on MS-COCO 2017 benchmark \cite{lin2014microsoft} (improving AP by 3.5 and 1.0 points compared to Hourglass \cite{newell2016stacked} and CPN \cite{chen2018cascaded} respectively, with the similar backbone and input size).

In the feature extractor, the deeper the layer is, the more specific the learned features are. For example, the first layer may learn overall features by abstracting the pixels and encoding the edges; the second layer may learn how to arrange the edges; the third layer encodes the face; the fourth layer encodes the eyes. Simply to see that the model needs to learn specialized features like eyes, nose because they correspond to the human keypoints. In particular, there are many cases of occluded keypoints. For example, the wrist is behind the back, so the wrist may not be detected. However, we actually can infer the wrist thanks to other keypoints such as elbow, shoulder, or even human skeleton. This means the model needs not only specific features but also overall patterns. 

This paper is inspired by the idea that the simple architecture could be ameliorated if it can learn the features from multiple resolutions, for the high resolution allows capturing overall information and the low resolution aims to extract specific characteristics. We propose novel network architectures utilizing the simple baseline \cite{xiao2018simple}, combining with the multi-resolution learning strategy. Our first approach achieves the multi-resolution heatmaps after the lowest-resolution feature maps are obtained. To do so, we branch off at each resolution of the heatmap generator and add extra layers for heatmap generation. In our second approach, the networks directly learn the heatmap generation at each resolution of the feature extractor. Our experiments were conducted on two common benchmarks for human pose estimation: MS-COCO \cite{lin2014microsoft} and MPII \cite{andriluka20142d}. On the COCO \textit{val2017} dataset, our best model gains AP by 0.6 points compared to SimpleBaseline \cite{xiao2018simple} which has a similar backbone and input size. On the MPII dataset, our best model achieves PCKh@0.5 of 89.8.

\noindent
\textbf{Contributions:} Our main contributions are:
\begin{itemize}[noitemsep,nolistsep]
    \item We introduce two novel approaches to achieve multi-resolution representation for both heatmap generation and feature map extraction.
    \item Our architectures are simple yet effective, and experiments show the superiority of our approaches over numerous methods. 
    \item Our approaches could be applied to other tasks that have the architecture of encoder (feature extractor) - decoder (specific tasks) such as image captioning and image segmentation.
\end{itemize}

\section{Human pose estimation using Deconvolutional layers as the Heatmap generator}

\begin{figure}[h!]
\centering
\includegraphics[width=3.2in]{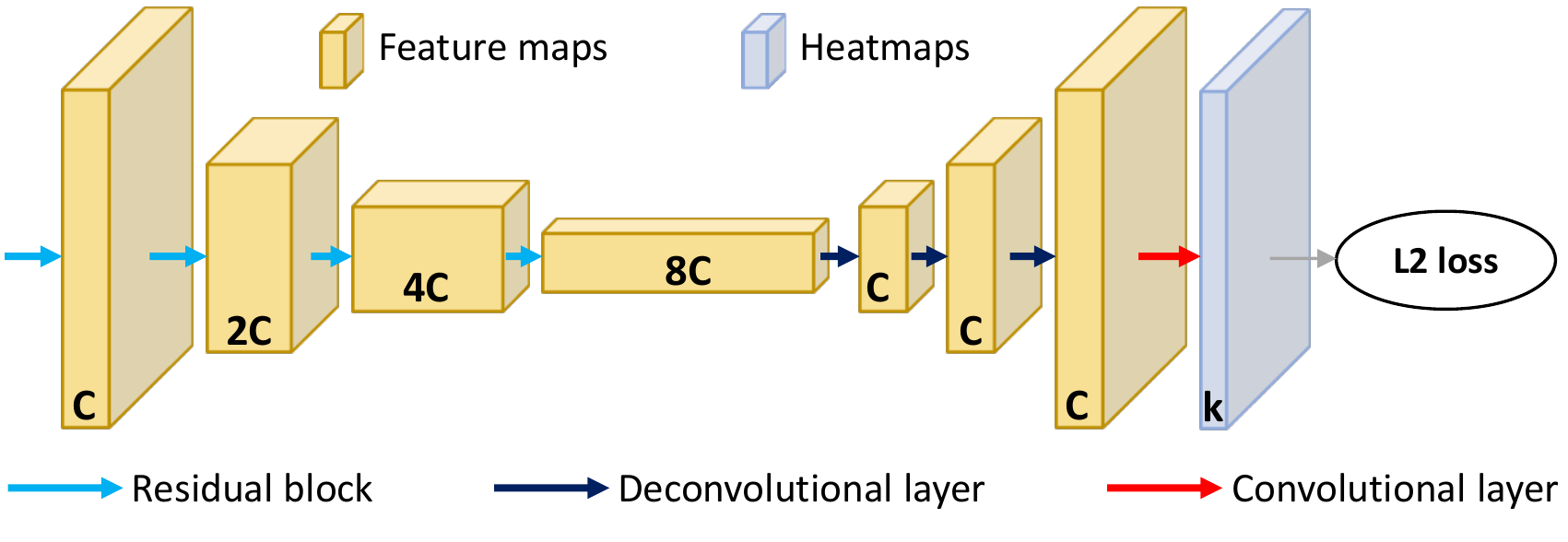}
\caption{Human pose estimation using deconvolutional layers as the heatmap generator.}
\label{fig-pose-resnet}
\end{figure}

This section presents the simple baseline \cite{xiao2018simple} whose the heatmap generator composed of deconvolutional layers. The network structure is illustrated in Fig. \ref{fig-pose-resnet}. From the input image, the model uses residual blocks to learn the features of the image. After each residual block, the resolution is decreased by half while the number of output channels is doubled. In Fig. \ref{fig-pose-resnet}, four residual blocks are working together as a feature extractor, and their numbers of output channels are $C$, $2C$, $4C$, and $8C$ respectively. We also use these notations for later architectures. 

After reaching $8C$ lowest-resolution feature maps, the network begins the top-down sequence of upsampling to obtain the high-resolution feature maps. Instead of using upsampling algorithms, SimpleBaseline \cite{xiao2018simple} leverages deconvolutional layers where each of them is built out of a transposed convolutional layer, 
a batch normalization, and a Relu activation. At last, a convolutional layer is added to generate $k$ high-resolution heatmaps representing the location confidence for all $k$ human keypoints. Mean Squared Error is used as the loss function between the predicted and ground-truth heatmaps:

\begin{equation}
    JointsLoss = \frac{\sum\limits_{i=1}^{k} (\frac{1}{w\times h}\sum\limits_{p=1}^{w}\sum\limits_{q=1}^{h}(H_{i,p,q}-\hat{H}_{i, p, q})^2)}{k},
\end{equation}

where $H_i$ and $\hat{H}_i$ are the ground-truth and predicted heatmap of the $i$th keypoint respectively, ($w$, $h$) is the size of the heatmap.

\section{Our method}

To investigate the impact of multi-resolution representation, in this section, we propose learning the multi-resolution representation for both the heatmap generator and the feature extractor. These two approaches are referred to as multi-resolution heatmap learning and multi-resolution feature map learning, respectively. We use ResNet 
as our feature extractor because it is the most common backbone network for image feature extraction. 

\subsection{Multi-resolution heatmap learning}

\begin{figure*}[h!]
\centering
\subfloat[MRHeatNet1]{\includegraphics[width=2.9in]{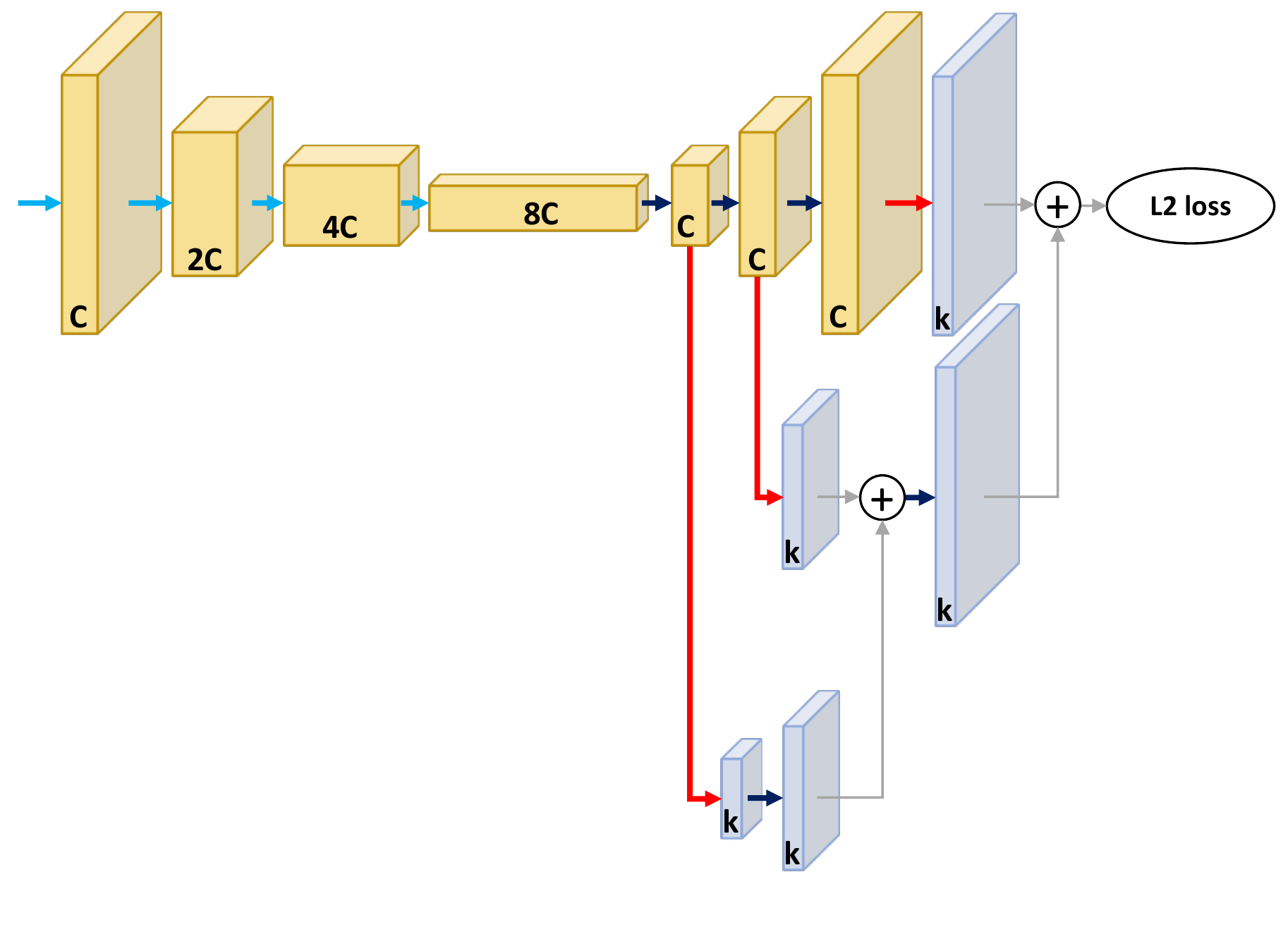}
\label{fig_multi_heatmap_1}}
\hfil
\subfloat[MRHeatNet2]{\includegraphics[width=2.9in]{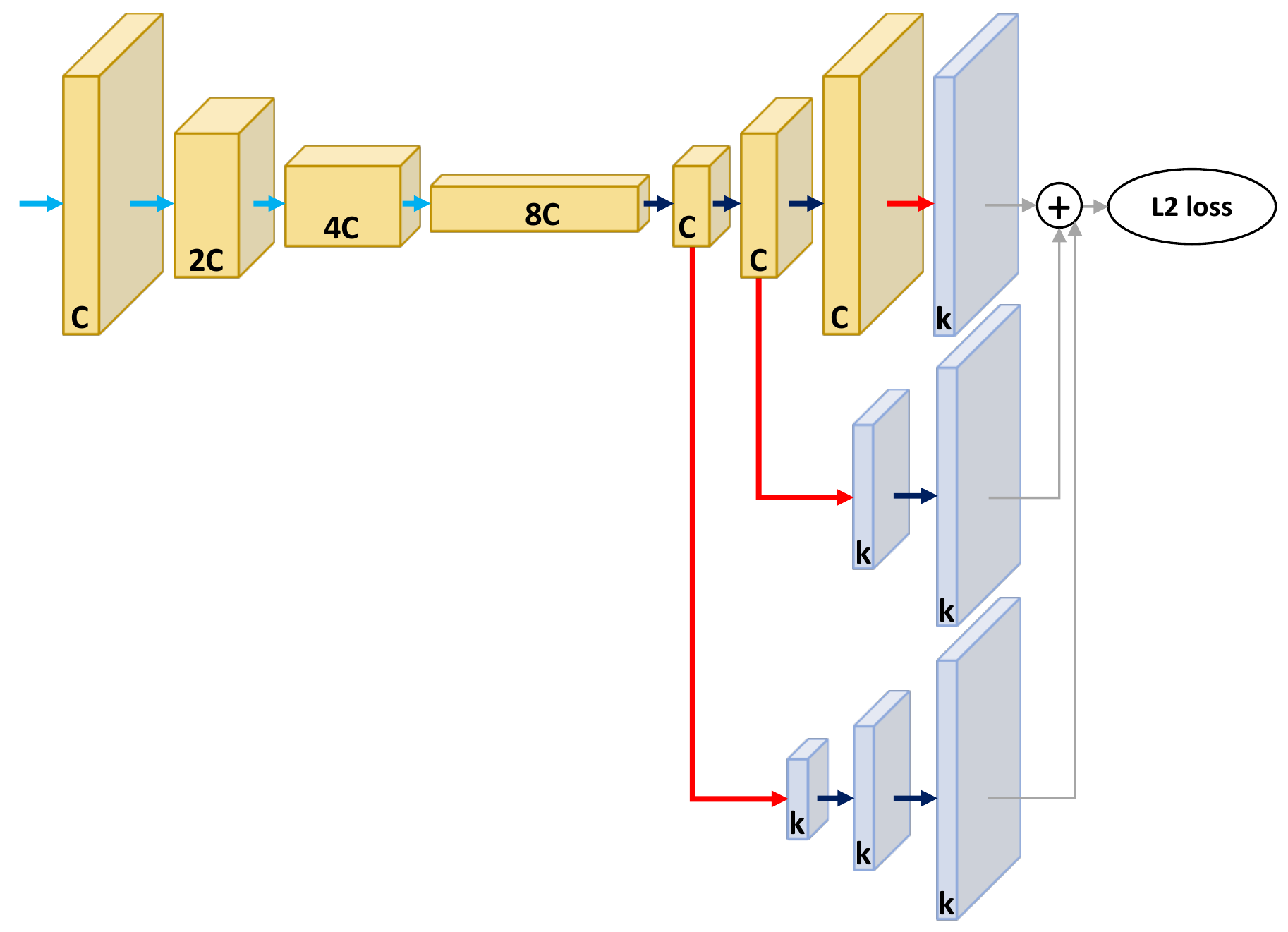}
\label{fig_multi_heatmap_2}}
\\
\includegraphics[width=7.1in]{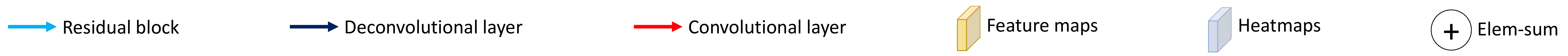}
\caption{Multi-resolution heatmap learning. We propose two architectures for generating the heatmaps at each resolution of the deconvolutional layers. (a) The lowest-resolution heatmaps are upsampled and then combined with the higher-resolution heatmaps. (b) The heatmaps at each resolution are individually learned and then combined at the end. The residual block halves the resolution of the input. The deconvolutional layer doubles the resolution of the input.}
\label{fig_multi-heatmap}
\end{figure*}

We started thinking about this kind of architecture by assuming that the ResNet backbone 
works very well on the image feature extraction. The architectures of the multi-resolution heatmap learning are illustrated in Fig. \ref{fig_multi-heatmap}. The lowest-resolution feature maps are fed into the sequence of deconvolutional layers to obtain the higher resolutions. The number of output channels of these deconvolutional layers is kept unchanged and is set to be equal to the number of output channels (denoted by $C$) of the first residual block.

In the baseline method, $k$ heatmaps are generated after obtaining the highest resolution. In our method, we branch off at each deconvolutional layer (excluding the highest-resolution deconvolutional layer) and add some convolutional layers to generate the low-resolution heatmaps. The higher-resolution heatmaps could be obtained from the low-resolution heatmaps by using extra deconvolutional layers. The reason we do so is that the high-resolution feature maps help generate the heatmaps with overall information while the low-resolution feature maps focus on specific characteristics. We propose two architectures with a slight difference as shown in Fig. \ref{fig_multi-heatmap}:

\begin{figure*}[h!]
\centering
\subfloat[MRFeaNet1]{\includegraphics[width=3.00in]{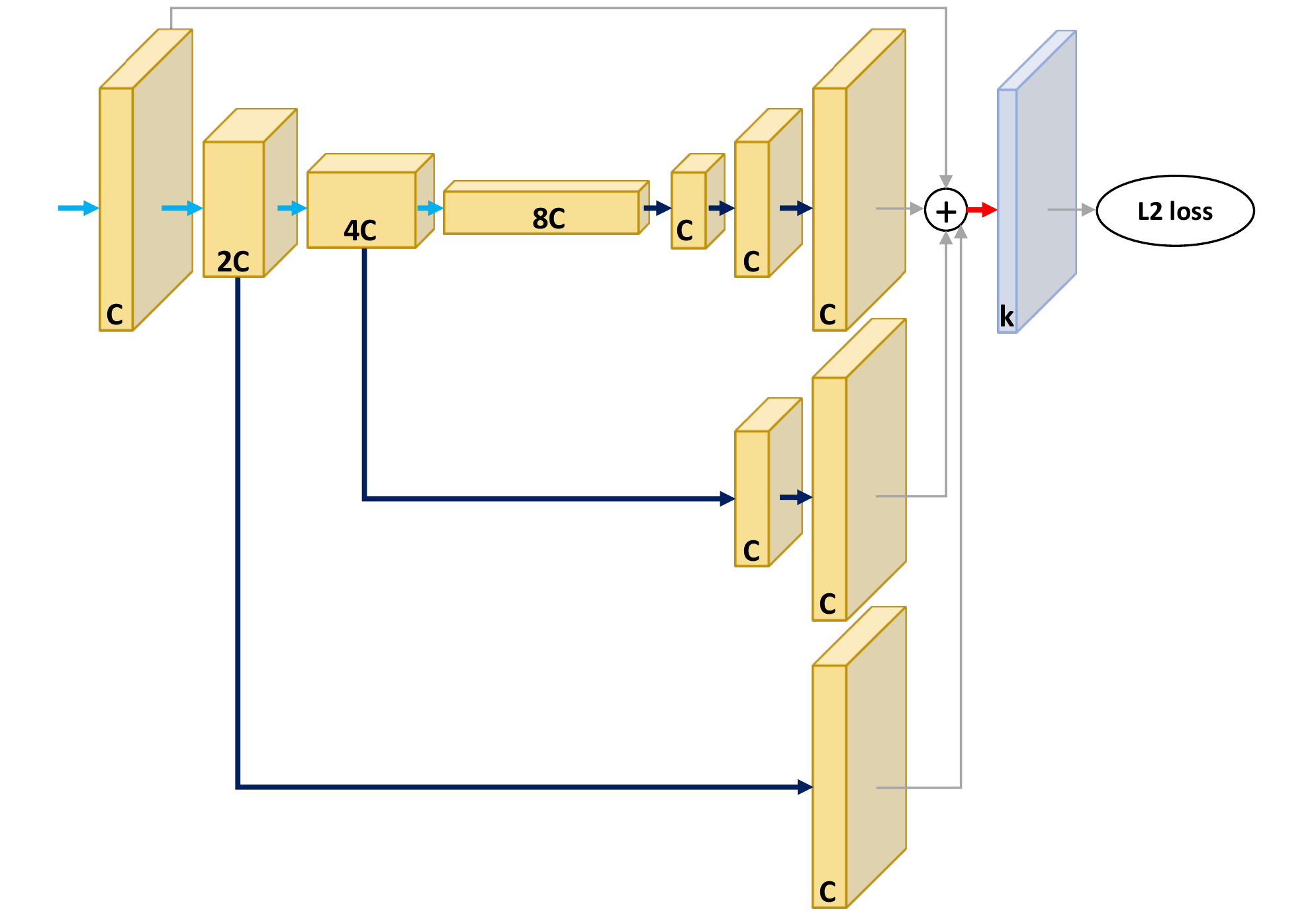}
\label{fig_hr_1}}
\hfil
\subfloat[MRFeaNet2]{\includegraphics[width=3.00in]{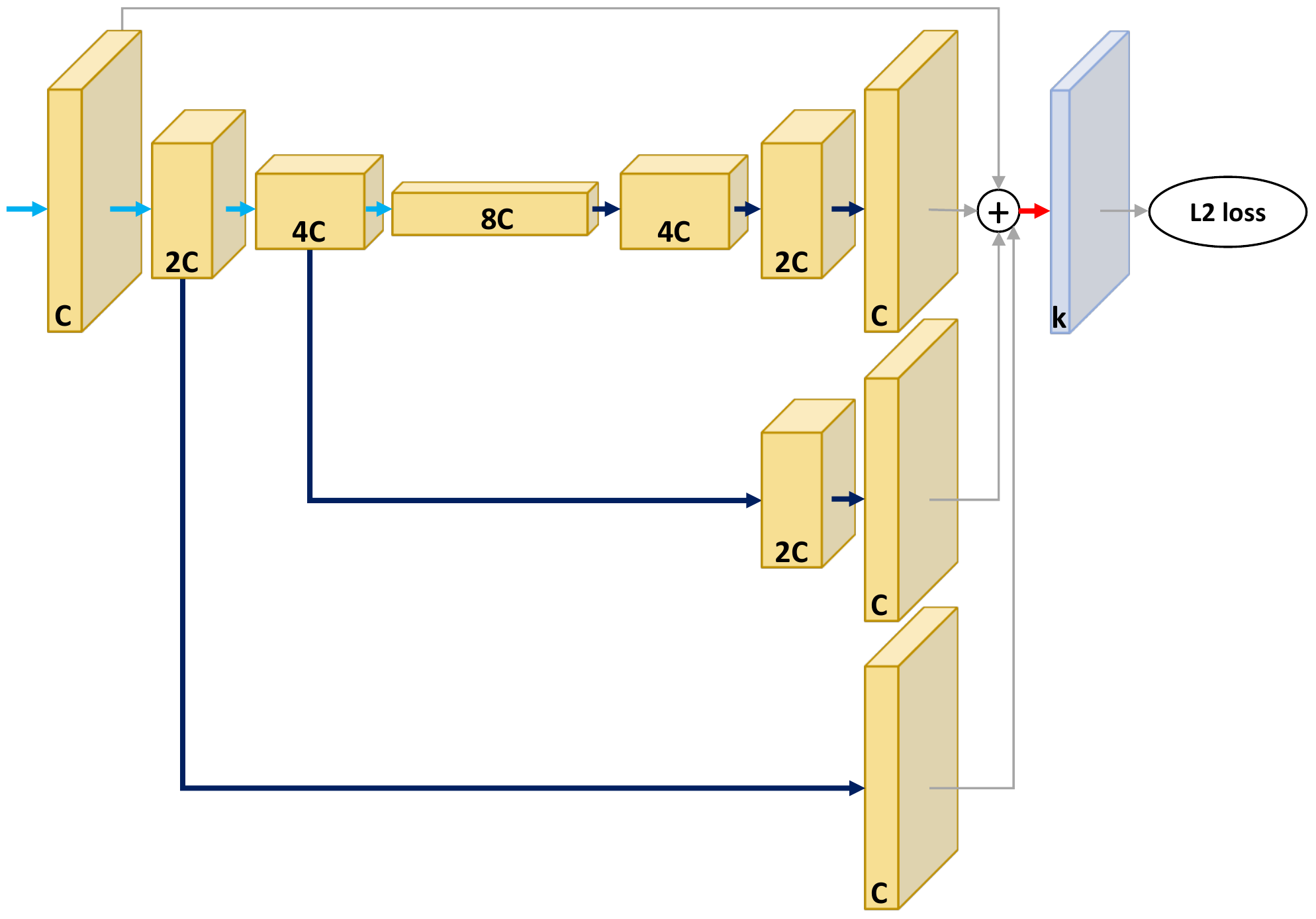}
\label{fig_hr_2}}
\caption{Multi-resolution feature map learning. We propose two architectures for learning the features at each resolution of the residual blocks. (a) The number of output channels of deconvolutional layers is kept unchanged. (b) The number of output channels is different among the deconvolutional layers. The highest-resolution heatmaps are obtained from the feature maps at each resolution of the feature extractor. Notations in Fig. \ref{fig_multi-heatmap} are also used here. The residual block halves the resolution of the input. The deconvolutional layer doubles the resolution of the input.}
\label{fig_hr}
\end{figure*}

\begin{itemize}
    \item In Fig. \ref{fig_multi_heatmap_1}, the lowest-resolution heatmaps are upsampled to the higher resolution (called medium resolution) and then combined with the heatmaps generated at this medium resolution. The result of this combination is fed into a deconvolutional layer to obtain the highest-resolution heatmaps.
    \item With a small change, in Fig. \ref{fig_multi_heatmap_2}, the heatmaps at each resolution are upsampled to the highest-resolution heatmaps independently and then combined at the end.
\end{itemize}

\subsection{Multi-resolution feature map learning}
\label{sec_multi_fea}

Instead of learning at each resolution of the heatmap generator as in the multi-resolution heatmap learning strategy, the multi-resolution feature map learning aims to directly learn how to generate the heatmaps at each resolution of the feature extractor (Fig. \ref{fig_hr}). At each residual block corresponding to each resolution of the feature extractor (excluding the lowest resolution), the network branches off and goes through respective deconvolutional layers to obtain the highest resolution. Especially, the branch from the highest-resolution residual block does not go through any deconvolutional layers but directly goes to the element-sum component. At last, a $1\times 1$ convolutional layer is added to generate $k$ predicted heatmaps for all $k$ keypoints.

Following this stream, we propose two architectures as illustrated in Fig. \ref{fig_hr_1} and Fig. \ref{fig_hr_2}. The main difference between these two architectures is the number of output channels of deconvolutional layers. In the network shown in Fig. \ref{fig_hr_1}, the number of output channels of all deconvolutional layers is set to be equal to the number of output channels (denoted by $C$) of the highest-resolution residual block, this may lead to an information loss.

The feature extractor consists of four residual blocks: the first residual block outputs $C$ feature maps with the size of $W\times H$, the second residual block aims to learn more features and outputs $2C$ feature maps with the size of $W/2\times H/2$, the third residual block outputs $4C$ feature maps with the size of $W/4\times H/4$, and the fourth residual block finally outputs $8C$ lowest-resolution feature maps with the size of $W/8\times H/8$. It is easy to see the principle of the image feature extraction here: the number of feature maps is increased by a factor of 2 (more features are learned) while the resolution is halved. Therefore, in the top-down sequence of upsampling, the resolution is increased two times, the number of feature maps should be decreased two times as well. For the network shown in Fig. \ref{fig_hr_1}, after the first deconvolutional layer in the main branch, the resolution of feature maps is increased two times, but the number of feature maps is decreased eight times (from $8C$ to $C$). Therefore, some previously learned information may be lost. To overcome this point, the architecture in Fig. \ref{fig_hr_2} uses the deconvolutional layers with the number of output channels depending on the number of feature maps extracted by the previously adjacent layer. For instance, after the fourth residual block, $8C$ lowest-resolution feature maps are outputted; as a result, the numbers of output channels of following deconvolutional layers are $4C$, $2C$, and $C$, respectively. The effectiveness of learning the heatmap generation from multiple resolutions of the feature extractor will be clarified in Section \ref{experiments}.


\section{Experiment}\label{experiments}
\paragraph{Dataset} 

We evaluate our architectures on two common benchmarks for human pose estimation: MS-COCO \cite{lin2014microsoft} and MPII \cite{andriluka20142d}. 

\begin{itemize}
    \item The COCO dataset contains more than 200k images and 250k person instances labeled with keypoints. Each person is annotated with 17 keypoints. We train our models on COCO \textit{train2017} dataset with 57k images and 150k person instances. Our models are evaluated on COCO \textit{val2017} and \textit{test-dev2017} dataset, with 5k and 20k images, respectively.
    \item The MPII dataset contains around 25k images with over 40k person samples. Each person is annotated with 16 joints. MPII covers 410 human activities collected from YouTube videos where the contents are everyday human activities. Since the annotations of MPII test set are not available, we train our models on a subset of 22k training samples and evaluate our models on a validation set of 3k samples \cite{tompson2015efficient}.
\end{itemize}

\paragraph{Evaluation metric} 

We use different metrics for our evaluation on the MS-COCO and MPII dataset:

\begin{itemize}
    \item In the COCO dataset, each person object has the ground-truth keypoints with the form $[x_1,y_1,v_1,...,x_k,y_k,v_k]$, where $x, y$ are the keypoint locations and $v$ is a visibility flag ($v=0$: not labeled, $v=1$: labeled but not visible, and $v=2$: labeled and visible). The standard evaluation metric is based on Object Keypoint Similarity (OKS):
    \begin{equation}
        OKS = \frac{\sum_{i} [exp(-d_{i}^{2}/2s^{2}k_{i}^{2})\delta(v_{i}>0)]}{\sum_{i} [\delta(v_{i}>0)]}
    \end{equation}
    In which, $d_{i}$ is the Euclidean distance between the detected and corresponding ground-truth keypoint, $v_{i}$ is the visibility flag of the ground-truth keypoint, $s$ is the object scale, and $k_{i}$ is a per-keypoint constant that controls falloff. Predicted keypoints that are not labeled ($v_{i}=0$) do not affect the OKS. The OKS plays the same role as the IoU in object detection, so the average precision (AP) and average recall (AR) scores could be computed if given the OKS.
    \item For the MPII dataset, we use Percentage of Correct Keypoints with respect to head (PCKh) metric \cite{andriluka20142d}. Firstly, we recall Percentage of Correct Keypoints (PCK) metric \cite{yang2011articulated}. PCK is the percentage of correct detection that falls within a tolerance range which is a fraction of torso diameter. The equation could be expressed as:
    \begin{equation}
        \frac{\left\lVert y_i - \hat{y}_i \right\rVert_2}{\left\lVert y_{rhip} - y_{lsho} \right\rVert_2}\leq r,
    \end{equation}
    where $y_i$ and $\hat{y}_i$ are the ground-truth and predicted location of the $i$th keypoint respectively, $y_{rhip}$ and $y_{lsho}$ are the ground-truth location of right hip and left shoulder respectively, r is the threshold bounded between 0 and 1. $\left\lVert y_{rhip} - y_{lsho} \right\rVert_2$ represents the torso diameter. For example, PCK@0.2 ($r=0.2$) means that: \textit{the distance between the predicted and ground-truth keypoint} $\leq$ $0.2$ $\times$ \textit{torso diameter}. PCKh is almost the same as PCK except that the tolerance range is a fraction of head size.
\end{itemize}

\paragraph{Network parameter} 

For all our experiments, we use ResNet 
as our backbone for the image feature extraction, consisting of 4 residual blocks as shown in Fig. \ref{fig_multi-heatmap} and Fig. \ref{fig_hr}. Each deconvolutional layer uses $4\times 4$ kernel filters. Each convolutional layer uses $1\times 1$ kernel filters. The numbers of output channels of the residual block, deconvolutional layer, and convolutional layer are denoted by $C$ and $k$ as shown in Fig. \ref{fig_multi-heatmap} and Fig. \ref{fig_hr}. $C$ is set to 256. $k$ is set to 17 or 16 for the COCO or MPII dataset respectively.

\begin{table*}[h!]
\centering
\caption{Comparisons on COCO \textit{val2017} dataset. OHKM means Online Hard Keypoints Mining \cite{chen2018cascaded}. Pretrain means the backbone is pre-trained on the ImageNet classification task.}
\label{table_coco}
\begin{tabular}{l|l|c|cccccccccc}
\hline
\multicolumn{1}{c|}{\textbf{Method}} & \multicolumn{1}{c|}{\textbf{Backbone}} & \textbf{Pretrain} & \textbf{AP}   & \textbf{AP$^{50}$} & \textbf{AP$^{75}$} & \textbf{AP$^{M}$}  & \textbf{AP$^{L}$}  & \textbf{AR}   & \textbf{AR$^{50}$} & \textbf{AR$^{75}$} & \textbf{AR$^{M}$}  & \textbf{AR$^{L}$}  \\ \hline
8-stage Hourglass \cite{newell2016stacked}                  & 8-stage Hourglass                     & N                 & 66.9          & -             & -             & -             & -             & -             & -             & -             & -             & -             \\
CPN \cite{chen2018cascaded}                                 & ResNet-50                             & Y                 & 68.6          & -             & -             & -             & -             & -             & -             & -             & -             & -             \\
CPN + OHKM \cite{chen2018cascaded}                          & ResNet-50                             & Y                 & 69.4          & -             & -             & -             & -             & -             & -             & -             & -             & -             \\
SimpleBaseline \cite{xiao2018simple}                     & ResNet-50                             & Y                 & 70.4          & 88.6          & 78.3 & 67.1          & 77.2          & 76.3          & 92.9          & 83.4          & 72.1          & 82.4          \\ \hline
MRHeatNet1                   & ResNet-50                             & Y                 & 70.2          & 88.5          & 77.6          & 66.8          & 77.2          & 76.2          & 92.8          & 83.0          & 71.8          & 82.4          \\
MRHeatNet2                    & ResNet-50                             & Y                 & 70.3          & 88.5          & 78.0          & 67.2 & 77.0          & 76.4          & 92.9          & 83.1          & 72.1          & 82.4          \\
MRFeaNet1                 & ResNet-50                             & Y                 & 70.6          & 88.7          & 78.1          & 67.3          & 77.5          & 76.5          & 92.9          & 83.3          & 72.1          & 82.7          \\ 
MRFeaNet2                 & ResNet-50                             & Y                 & 70.9 & 88.8 & 78.3 & 67.2 & 78.1 & 76.8 & 93.0 & 83.6 & 72.2 & 83.4 \\  \hline
SimpleBaseline \cite{xiao2018simple}                     & ResNet-101                             & Y                 & 71.4          & 89.3          & 79.3 & 68.1          & 78.1          & 77.1          & 93.4          & 84.0          & 73.0          & 83.2          \\
MRFeaNet2                 & ResNet-101                             & Y                 & 71.8 & 89.1 & 79.6 & 68.5 & 78.8 & 77.8 & \textbf{93.5} & 84.5 & 73.5 & 84.0 \\  \hline
SimpleBaseline \cite{xiao2018simple}                     & ResNet-152                             & Y                 & 72.0          & 89.3          & 79.8 & 68.7          & 78.9          & 77.8          & 93.4          & 84.6          & 73.6          & 83.9          \\
MRFeaNet2                 & ResNet-152                             & Y                 & \textbf{72.6} & \textbf{89.4} & \textbf{80.4} & \textbf{69.4} & \textbf{79.3} & \textbf{78.2} & 93.4 & \textbf{85.2} & \textbf{74.1} & \textbf{84.2} \\  \hline
\end{tabular}
\end{table*}

\begin{table*}[h!]
\centering
\caption{Comparisons on COCO \textit{test-dev} dataset.}
\label{table_coco-testdev}
\begin{tabular}{l|l|c|cccccccccc}
\hline
\multicolumn{1}{c|}{\textbf{Method}} & \multicolumn{1}{c|}{\textbf{Backbone}} & \textbf{Input size} & \textbf{AP}   & \textbf{AP$^{50}$} & \textbf{AP$^{75}$} & \textbf{AP$^{M}$}  & \textbf{AP$^{L}$}  & \textbf{AR}   & \textbf{AR$^{50}$} & \textbf{AR$^{75}$} & \textbf{AR$^{M}$}  & \textbf{AR$^{L}$}  \\ \hline
\multicolumn{13}{c}{Bottom-up approach: keypoint detection and grouping}                                                                                                                                                                                          \\ \hline
OpenPose \cite{cao2017realtime}                            & \multicolumn{1}{c|}{-}                 & -                   & 61.8          & 84.9          & 67.5          & 57.1          & 68.2          & -          & -             & -             & -             & -             \\ 
Associative Embedding \cite{newell2017associative}              & \multicolumn{1}{c|}{-}                 & -                   & 65.5          & 86.8          & 72.3          & 60.6          & 72.6          & 70.2          & 89.5             & 76.0             & 64.6             & 78.1             \\ 
PersonLab \cite{papandreou2018personlab}                          & ResNet-152                 & -                   & 68.7          & 89.0          & 75.4          & 64.1          & 75.5          & 75.4          & 92.7             & 81.2             & 69.7             & \textbf{83.0}             \\ 
MultiPoseNet \cite{kocabas2018multiposenet}                       & \multicolumn{1}{c|}{-}                 & -                   & 69.6          & 86.3          & 76.6          & 65.0          & 76.3 & 73.5          & 88.1             & 79.5             &  68.6             & 80.3             \\ \hline
\multicolumn{13}{c}{Top-down approach: person detection and single-person keypoint detection}                                                                                                                                                                     \\ \hline
Mask-RCNN \cite{he2017mask}                          & ResNet-50-FPN                         & -                   & 63.1          & 87.3          & 68.7          & 57.8          & 71.4          & -             & -             & -             & -             & -             \\ 
G-RMI \cite{papandreou2017towards}                              & ResNet-101                            & 353 $\times$ 257           & 64.9          & 85.5          & 71.3          & 62.3          & 70.0          & 69.7          & 88.7             & 75.5             & 64.4             & 77.1             \\ 
Integral Pose Regression \cite{sun2018integral}            & ResNet-101                            & 256 $\times$ 256           & 67.8          & 88.2          & 74.8          & 63.9          & 74.0          & -             & -             & -             & -             & -             \\ 
G-RMI + extra data \cite{papandreou2017towards}                 & ResNet-101                            & 353 $\times$ 257           & 68.5          & 87.1          & 75.5          & 65.8          & 73.3          & 73.3          & 90.1             & 79.5             & 68.1             & 80.4             \\ 
SimpleBaseline \cite{xiao2018simple}                      & ResNet-50                             & 256 $\times$ 192           & 70.0          & 90.9 & 77.9          & 66.8          & 75.8          & 75.6          & 94.5          & 83.0          & 71.5          & 81.3          \\
SimpleBaseline \cite{xiao2018simple}                      & ResNet-101                             & 256 $\times$ 192           & 70.9          & 91.1 & 79.3          & 67.9          & 76.7          & 76.7          & \textbf{94.9}          & 84.2          & 72.7          & 82.2          \\
SimpleBaseline \cite{xiao2018simple}                      & ResNet-152                             & 256 $\times$ 192           & 71.6          & \textbf{91.2} & \textbf{80.1}          & 68.7          & 77.2          & 77.2          & \textbf{94.9}          & \textbf{85.0}          & 73.4          & 82.6          \\ \hline
\multicolumn{13}{c}{Our multi-resolution representation learning models}                                                                                                                                                                                          \\ \hline
MRHeatNet1                    & ResNet-50                             & 256 $\times$ 192           & 69.7          & 90.8          & 77.8          & 66.6          & 75.4          & 75.4          & 94.4          & 82.9          & 71.3          & 81.1          \\ 
MRHeatNet2                    & ResNet-50                             & 256 $\times$ 192           & 69.9          & 90.8          & 78.3          & 66.9          & 75.6          & 75.6          & 94.5          & 83.3          & 71.6          & 81.2          \\ 
MRFeaNet1                     & ResNet-50                             & 256 $\times$ 192           & 70.1          & 90.7          & 78.4          & 67.0          & 75.9          & 75.8          & 94.3          & 83.3          & 71.7          & 81.3          \\ 
MRFeaNet2                     & ResNet-50                             & 256 $\times$ 192           & 70.4 & 90.9 & 78.7 & 67.3 & 76.3 & 76.2 & 94.6 & 83.7 & 72.0 & 81.9 \\
MRFeaNet2                     & ResNet-101                             & 256 $\times$ 192           & 71.2 & 91.0 & 79.6 & 68.2 & 76.9 & 77.0 & 94.7 & 84.5 & 72.9 & 82.5 \\
MRFeaNet2                     & ResNet-152                             & 256 $\times$ 192           & \textbf{71.8} & \textbf{91.2} & \textbf{80.1} & \textbf{68.9} & \textbf{77.5} & \textbf{77.4} & 94.8 & 84.9 & \textbf{73.5} & 82.8 \\ \hline
\end{tabular}
\end{table*}

\subsection{Experimental results on COCO dataset}
\label{sec_exp_coco}

\textbf{Training}. The data pre-processing and augmentation follow the setting in \cite{xiao2018simple}. The ground-truth human bounding box is extended in height or width to a fixed aspect ratio ($height : width = 4 : 3$). The human box after cropped from the image is resized to a fixed size of $256\times 192$ for a fair comparison with \cite{newell2016stacked, chen2018cascaded, xiao2018simple}. The data augmentation includes random rotation ($\pm30\degree$), random scale ($\pm40\%$), and flip. We use Adam optimizer. 
The batch size is 64. The learning schedule is set up as follows: the base learning rate is set to $1e-3$, and is dropped to $1e-4$ and $1e-5$ at the $120$th and $150$th epoch, respectively. The training process is terminated within 170 epochs.

\textbf{Testing}. We use the two-stage top-down paradigm, similar to \cite{chen2018cascaded, xiao2018simple}. Keypoint locations are obtained by using the highest heatvalue's location in predicted heatmaps and a quarter offset in the direction from the highest response to the second-highest response.

\textbf{Comparisons on COCO \textit{val2017} dataset}. TABLE \ref{table_coco} reports our evaluation results compared to Hourglass \cite{newell2016stacked}, CPN \cite{chen2018cascaded}, and SimpleBaseline \cite{xiao2018simple}. Note that the results of Hourglass \cite{newell2016stacked} are cited from \cite{chen2018cascaded}. For the fair comparison, we use the faster-RCNN detector \cite{ren2015faster} with the detection AP of 56.4 (being the same with that of SimpleBaseline \cite{xiao2018simple}) while the person detection AP of Hourglass \cite{newell2016stacked} and CPN \cite{chen2018cascaded} is 55.3.

As shown in TABLE \ref{table_coco}, both our architectures outperform Hourglass \cite{newell2016stacked} and CPN \cite{chen2018cascaded}. With the same ResNet-50 backbone, our MRFeaNet2 achieves an AP score of 70.9, improving the AP by 4.0 and 2.3 points compared to Hourglass and CPN respectively. Online Hard Keypoints Mining (OHKM) proved the efficiency when helping CPN gain the AP by 0.8 points (from 68.6 to 69.4), but still being 1.5 points lower than the AP of MRFeaNet2.

Compared to SimpleBaseline \cite{xiao2018simple}, our multi-resolution heatmap learning architectures have slightly worse performance. In the case of using the ResNet-50 backbone, SimpleBaseline has the AP score of 70.4 while the AP scores of MRHeatNet1 and MRHeatNet2 are 70.2 and 70.3 respectively. This may be explained that the deconvolutional layers cannot completely recover all information which the feature extractor already learned, so only learning from the outputs of deconvolutional layers is not enough to generate the heatmaps.

On the other hand, our multi-resolution feature map learning architectures have better performance compared to SimpleBaseline \cite{xiao2018simple}. With the ResNet-50 backbone, MRFeaNet1 gains AP by 0.2 points while the AP of MRFeaNet2 increases by 0.5 points. MRFeaNet2 still obtains the AP improvement of 0.4 and 0.6 points compared to SimpleBaseline in the case of using the ResNet-101 and ResNet-152 backbone, respectively. This proves that learning heatmap generation from multiple resolutions of the feature extractor can help improve the performance of keypoint prediction.

\textbf{Comparisons on COCO \textit{test-dev} dataset}. TABLE \ref{table_coco-testdev} shows the performance of our models and previous methods on the COCO \textit{test-dev} dataset. Note that the results of SimpleBasline \cite{xiao2018simple} are reproduced by us using the provided models. We use the human detector with the person detection AP of 60.9 on COCO \textit{test-dev} for SimpleBasline and our models. Our networks outperform bottom-up approaches. Our MRFeaNet2 achieves the AP improvement of 2.2 points compared to MultiPoseNet \cite{kocabas2018multiposenet}. In comparison with top-down approaches, our models are better even with the smaller backbone and image size. Our MRFeaNet2, which uses the ResNet-50 backbone, obtains the AP of 70.4 while the AP score of G-RMI \cite{papandreou2017towards} is 68.5 even using the larger backbone network, larger image size, and extra training data. Compared to SimpleBaseline \cite{xiao2018simple}, our MRFeaNet2 still improves the AP by 0.4, 0.3, and 0.2 points in the case of using the ResNet-50, ResNet-101, and ResNet-152 backbone, respectively.

\subsection{Experimental results on MPII dataset}

\textbf{Training}. The data pre-processing and augmentation are similar to the setting in the experiment on the COCO dataset. The input size of human bounding box is set to $256\times 256$ for a fair comparison with other methods. The data augmentation includes random rotation ($\pm30\degree$), random scale ($\pm25\%$), and flip. Adam optimizer 
is also used. The batch size is 64. The learning rate starts from $1e-3$, drops to $1e-4$ and $1e-5$ at the $90$th and $120$th epoch, respectively. The training process is terminated within $140$ epochs.

\begin{table}[h!]
\centering
\caption{Comparisons on MPII dataset (PCKh@0.5). ($^{50}$), ($^{101}$), or ($^{152}$) means the ResNet-50, ResNet-101, or ResNet-152 backbone is used, respectively.}
\label{table_mpii}
\begin{tabular}{@{\hskip 0.03in}l@{\hskip 0.04in}|@{\hskip 0.05in}c@{\hskip 0.1in}c@{\hskip 0.1in}c@{\hskip 0.1in}c@{\hskip 0.1in}c@{\hskip 0.1in}c@{\hskip 0.1in}c@{\hskip 0.05in}|@{\hskip 0.03in}c@{\hskip 0.03in}}
\hline
\multicolumn{1}{c|@{\hskip 0.05in}}{\textbf{Method}} & \textbf{Hea} & \textbf{Sho} & \textbf{Elb} & \textbf{Wri} & \textbf{Hip}  & \textbf{Kne} & \textbf{Ank} & \textbf{Total} \\ \hline
Pishchulin et al. \cite{pishchulin2013strong}                   & 74.3          & 49.0              & 40.8           & 34.1           & 36.5          & 34.4          & 35.2           & 44.1          \\
Tompson et al. \cite{tompson2014joint}                      & 95.8          & 90.3              & 80.5           & 74.3           & 77.6          & 69.7          & 62.8           & 79.6          \\
Carreira et al. \cite{carreira2016human}                     & 95.7          & 91.7              & 81.7           & 72.4           & 82.8          & 73.2          & 66.4           & 81.3          \\
Tompson et al. \cite{tompson2015efficient}                     & 96.1          & 91.9              & 83.9           & 77.8           & 80.9          & 72.3          & 64.8           & 82.0          \\
Hu et al. \cite{hu2016bottom}                          & 95.0          & 91.6              & 83.0           & 76.6           & 81.9          & 74.5          & 69.5           & 82.4          \\
Pishchulin et al. \cite{pishchulin2016deepcut}                  & 94.1          & 90.2              & 83.4           & 77.3           & 82.6          & 75.7          & 68.6           & 82.4          \\
Lifshitz et al. \cite{lifshitz2016human}                    & \textbf{97.8} & 93.3              & 85.7           & 80.4           & 85.3          & 76.6          & 70.2           & 85.0          \\
Gkioxary et al. \cite{gkioxari2016chained}                    & 96.2          & 93.1              & 86.7           & 82.1           & 85.2          & 81.4          & 74.1           & 86.1          \\
Rafi et al. \cite{rafi2016efficient}                        & 97.2          & 93.9              & 86.4           & 81.3           & 86.8          & 80.6          & 73.4           & 86.3          \\
Belagiannis et al. \cite{belagiannis2017recurrent}                 & 97.7          & 95.0              & 88.2           & 83.0           & 87.9          & 82.6          & 78.4           & 88.1          \\
Insafutdinov et al. \cite{insafutdinov2016deepercut}                & 96.8          & 95.2              & 89.3           & 84.4  & 88.4          & 83.4          & 78.0           & 88.5          \\
Wei et al. \cite{wei2016convolutional}                         & \textbf{97.8} & 95.0              & 88.7           & 84.0           & 88.4          & 82.8          & 79.4           & 88.5          \\ \hline
SimpleBaseline$^{50}$ \cite{xiao2018simple}                   & 96.4          & 95.3              & 89.0           & 83.2           & 88.4          & 84.0          & 79.6           & 88.5          \\
MRHeatNet1$^{50}$                   & 96.7          & 95.2              & 88.9           & 83.8           & 88.1          & 83.6          & 78.6           & 88.4          \\ 
MRHeatNet2$^{50}$                   & 96.8          & 95.5     & 88.6           & 83.8           & 88.5          & 83.6          & 78.7           & 88.5          \\
MRFeaNet1$^{50}$                & 96.5          & 95.5     & 89.6  & 84.3           & 88.6 & 84.6 & 80.6           & 89.1 \\ 
MRFeaNet2$^{50}$                & 96.6          & 95.4              & 88.9           & 83.9           & 88.5          & 84.6 & 80.9  & 88.9          \\ \hline
SimpleBaseline$^{101}$ \cite{xiao2018simple}                   & 96.9          & 95.9              & 89.5           & 84.4           & 88.4          & 84.5          & 80.7           & 89.1          \\
MRHeatNet1$^{101}$                   & 96.7          & 95.7              & 89.7           & 84.4           & 89.1          & 84.7          & 81.4           & 89.3          \\ 
MRHeatNet2$^{101}$                   & 97.4          & 95.6     & 89.3           & 84.2           & 89.0          & 84.9          & 81.2           & 89.3          \\
MRFeaNet1$^{101}$                & 96.8          & 95.6     & 89.4  & 84.6           & 89.2 & 85.2 & 81.2           & 89.4 \\ 
MRFeaNet2$^{101}$                & 96.6          & 95.2              & 89.3           & 84.2           & 89.2          & 85.9 & 81.6  & 89.3          \\ \hline
SimpleBaseline$^{152}$ \cite{xiao2018simple}                   & 97.0          & 95.9              & 90.0           & 85.0           & 89.2          & 85.3          & 81.3           & 89.6          \\
MRHeatNet1$^{152}$                   & 96.8          & \textbf{96.0}              & 90.1           & 84.4           & 88.9          & 85.3          & 81.4           & 89.5          \\
MRHeatNet2$^{152}$                   & 96.9          & 95.6     & 89.9           & 84.6           & 88.9          & \textbf{86.0}          & 81.2           & 89.5          \\
MRFeaNet1$^{152}$                & 97.2          & 95.9     & \textbf{90.2}  & \textbf{85.3}           & \textbf{89.3} & 85.4 & \textbf{82.0}           & \textbf{89.8} \\
MRFeaNet2$^{152}$                & 96.7          & 95.4              & 89.9           & 85.1           & 88.8          & 85.7 & 81.8  & 89.5          \\ \hline
\end{tabular}
\end{table}

\begin{figure}[h!]
\centering
\includegraphics[width=3.2in]{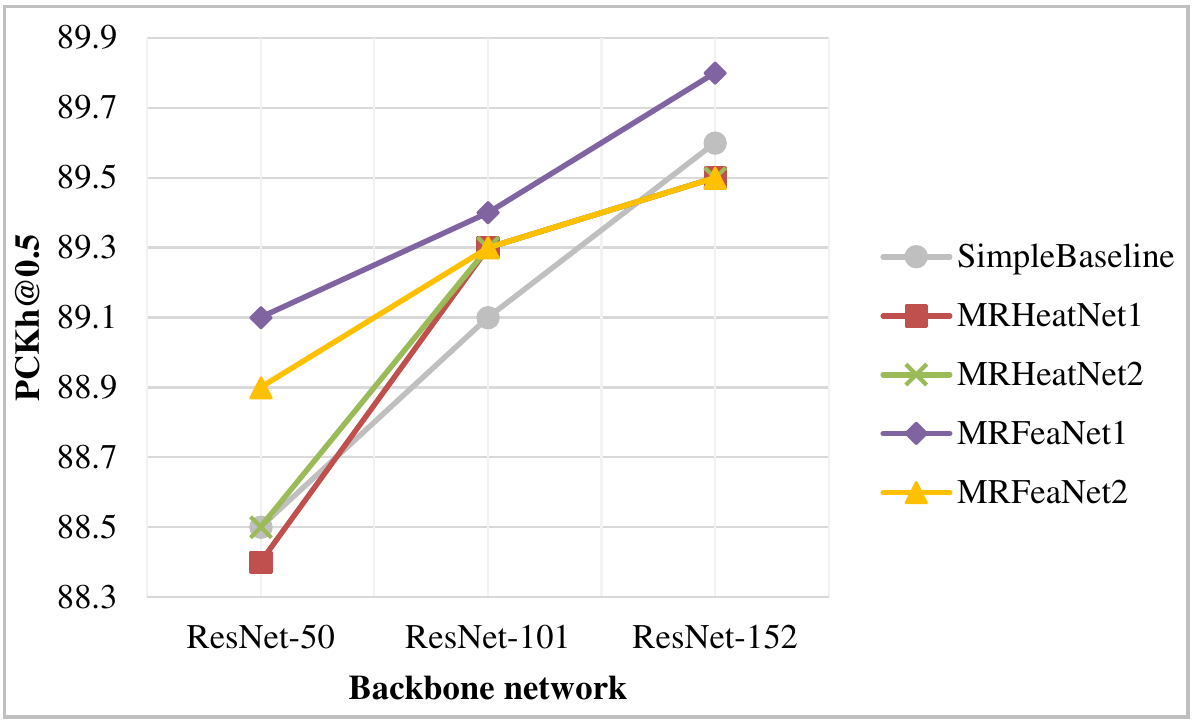}  
\caption{PCKh@0.5 score of SimpleBaseline and our models on MPII dataset.}
\label{fig_mpii_graph}
\end{figure}

\begin{figure}[h!]
\centering
\includegraphics[width=3.1in]{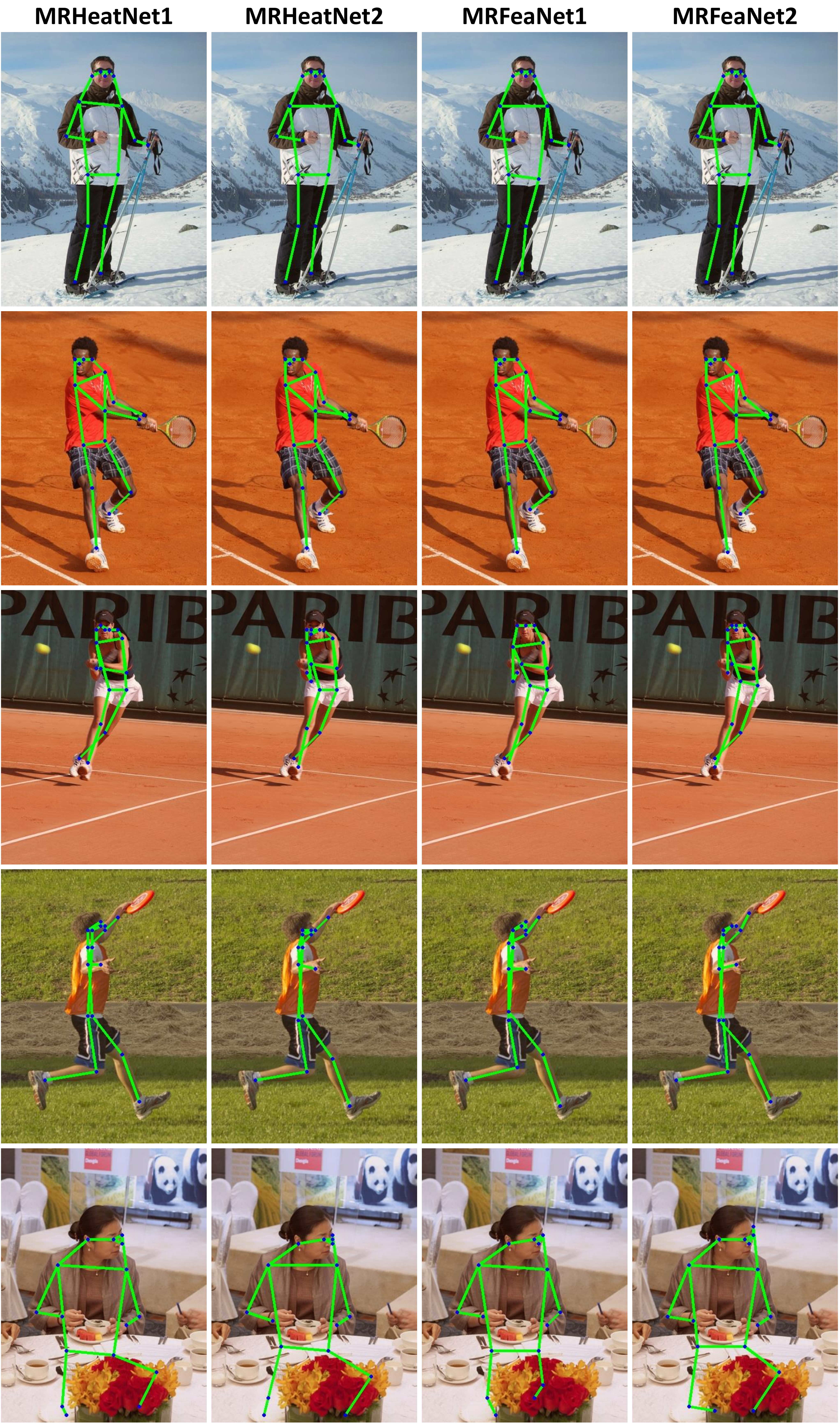}  
\caption{Qualitative results of our proposed architectures on COCO \textit{test2017} dataset.}
\label{fig_qualitative}
\end{figure}

\begin{figure*}[h!]
\centering
\includegraphics[width=5.5975in]{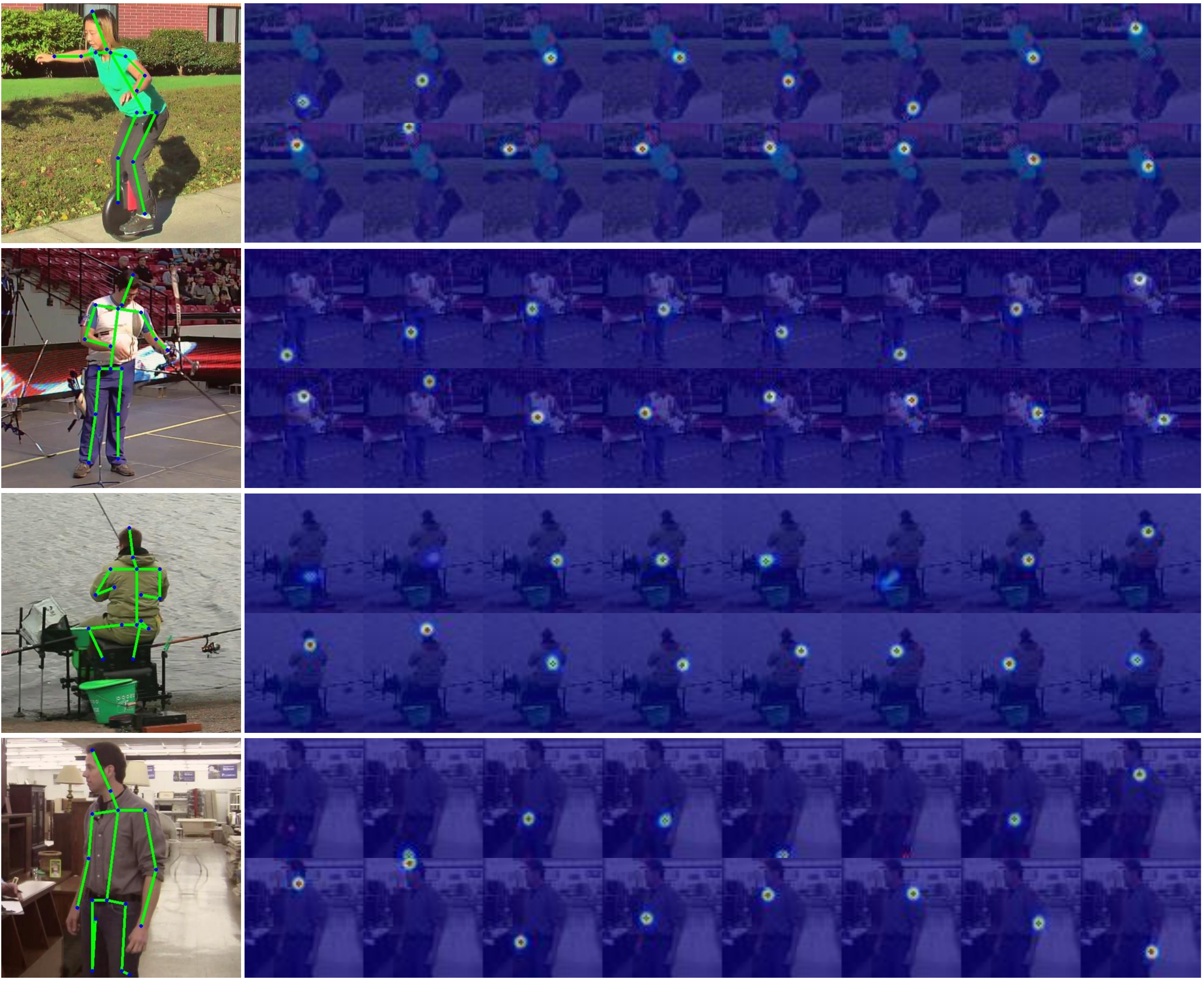}
\caption{Qualitative results of our MRFeaNet1$^{152}$ on MPII test set. Each prediction has 16 heatmaps corresponding to 16 human keypoints. From left to right, top to bottom, these 16 keypoints are right ankle, right knee, right hip, left hip, left knee, left ankle, pelvis, thorax, upper neck, head top, right wrist, right elbow, right shoulder, left shoulder, left elbow, and left wrist.}
\label{fig_qualitative_mpii}
\end{figure*}

\textbf{Testing}. We use the human bounding boxes provided with the images. TABLE \ref{table_mpii} shows the PCKh scores of our architectures and previous methods at $r=0.5$. The results of SimpleBaseline \cite{xiao2018simple} are reproduced by us using the provided models.

Similar to the experiments on the COCO dataset, our multi-resolution representation learning architectures outperform numerous previous methods. In comparison with SimpleBaseline \cite{xiao2018simple}, the multi-resolution feature map learning method achieves better performance. Our MRFeaNet1 gains the PCKh@0.5 score by 0.6, 0.3 and 0.2 points compared to SimpleBaseline in the case of using the ResNet-50, ResNet-101, and ResNet-152 backbone, respectively.

On the other hand, the results also show that the performance could be improved if using the larger backbone network. To make this statement clear, the PCKh@0.5 scores of SimpleBaseline \cite{xiao2018simple} and our models are presented on a chart as shown in Fig. \ref{fig_mpii_graph}. MRFeaNet1$^{152}$, which is the best model on the MPII dataset, obtains the score improvement of 0.4 and 0.7 points compared to MRFeaNet1$^{101}$ and MRFeaNet1$^{50}$ respectively. MRHeatNet1 achieves the highest improvement which is 1.1 points when the backbone network is transformed from ResNet-50 to ResNet-152.

MRFeaNet2 is superior on the COCO dataset (Section \ref{sec_exp_coco}) while MRFeaNet1 is dominant on the MPII dataset. This performance difference might relate to the architecture difference between these two networks. MRFeaNet2 has the number of output channels different among the deconvolutional layers, making it more robust and able to handle more complex contents compared to MRFeaNet1, as explained in Section \ref{sec_multi_fea}. The COCO dataset consists of 17 keypoints while the MPII dataset has 16 keypoints, so the COCO dataset is more complex than the MPII dataset. As a result, MRFeaNet2 is superior compared to MRFeaNet1 on the COCO dataset, and MRFeaNet1 plays the efficiency in processing the less complex skeleton as in the MPII dataset.

\subsection{Qualitative results}

\textbf{Qualitative results on COCO \textit{test2017} dataset}. We use our models trained on the COCO \textit{train2017} dataset with the ResNet-50 backbone to visualize human keypoint prediction. Our qualitative results on the unseen images of the COCO \textit{test2017} dataset are shown as in Fig. \ref{fig_qualitative}. Both our models work well on the simple cases (the 1$^{st}$ and 2$^{nd}$ row).

\begin{itemize}
    \item The figures in the 3$^{rd}$ and 4$^{th}$ row are harder with some occluded keypoints, but the multi-resolution feature map learning models still relatively precisely predict the human keypoints. The multi-resolution heatmap learning models do not work well: MRHeatNet1 omits the right elbow in the 3$^{rd}$ row, and the eye detection of MRHeatNet2 is not reasonable in both of these two cases. 
    \item In the 5$^{th}$ row, both legs of the woman are hidden under the table, but both of our models can make their opinion. The prediction results are different among the models. If carefully looking at the hip prediction, the locations proposed by MRFeaNet2 are the most reasonable result.
\end{itemize}

\textbf{Qualitative results on MPII dataset}. We use our MRFeaNet1 model trained on a subset of the MPII training set with the ResNet-152 backbone to visualize human keypoint prediction. Fig. \ref{fig_qualitative_mpii} shows the keypoint predictions and corresponding heatmaps on the unseen images of the MPII test set. Each heatmap represents the location confidence of the respective keypoint. With the simple cases as in the 1$^{st}$ and 2$^{nd}$ row, all keypoints are predicted with high confidence.

\begin{itemize}
    \item The man in the 3$^{rd}$ row has his right leg and left ankle occluded, so the prediction of these keypoints has low confidence. However, all prediction results of this case are reasonable and acceptable.
    \item Especially, the man in the 4$^{th}$ row has two ankles not displayed, so the ankle prediction is unreasonable. The heatmaps corresponding to these two ankles are suitable and meaningful, where there is no location predicted with high confidence.
\end{itemize}

\section{Conclusion}

In this paper, we introduce two novel approaches for multi-resolution representation learning solving human pose estimation. The first approach reconciles a multi-resolution representation learning strategy with the heatmap generator where the heatmaps are generated at each resolution of the deconvolutional layers. The second approach achieves the heatmap generation from each resolution of the feature extractor. While our multi-resolution feature map learning models outperform the baseline and many previous methods, the proposed architectures are relatively straightforward and integrable. The future work includes the applications to other tasks that have the architecture of encoder-decoder (feature extraction - specific tasks) such as image captioning and image segmentation.

\section*{Acknowledgment}
This work was supported by the Korea-EU Joint Research Support Project of the National Research Foundation of Korea (NRF) funded by the Ministry of Science, ICT and Future Planning of Korea (NRF-2016K1A3A7A0395205414); the MSIT(Ministry of Science and ICT), Korea, under the Grand Information Technology Research Center support program(IITP-2020-0-01489) supervised by the IITP(Institute for Information \& communications Technology Planning \& Evaluation); and the Technology Innovation Program(or Industrial Strategic Technology development Program, 2000682, Development of Automated Driving Systems and Evaluation) funded by the Ministry of Trade, Industry \& Energy(MOTIE,Korea).




%
\bibliography{example_paper}
\bibliographystyle{IEEEtran}

\end{document}